\documentclass[10pt]{article}
\usepackage[preprint]{tmlr}
\usepackage[utf8]{inputenc}
\usepackage[T1]{fontenc}
\usepackage{microtype}
\usepackage{amsmath}
\usepackage{amsfonts}
\usepackage{amssymb}
\usepackage{tikz}
\usetikzlibrary{positioning, fit, arrows.meta, backgrounds, calc}
\usepackage{graphicx}
\usepackage{booktabs}
\usepackage{float}
\usepackage[ruled,vlined,linesnumbered]{algorithm2e}
\usepackage{enumitem}
\usepackage{tabularx}
\usepackage{multirow}
\usepackage{url}
\usepackage{placeins}
\usepackage[colorlinks=true, citecolor=blue, linkcolor=blue, urlcolor=blue]{hyperref}
\usepackage{xcolor}

\definecolor{faGreen}{HTML}{D5E8D4}
\definecolor{faGreenLine}{HTML}{82B366}
\definecolor{faYellow}{HTML}{FFF2CC}
\definecolor{faYellowLine}{HTML}{D6B656}
\definecolor{faBlue}{HTML}{DAE8FC}
\definecolor{faBlueLine}{HTML}{6C8EBF}
\definecolor{faRed}{HTML}{F8CECC}
\definecolor{faRedLine}{HTML}{B85450}
\definecolor{faPurple}{HTML}{E1D5E7}
\definecolor{faPurpleLine}{HTML}{9673A6}
\definecolor{faGray}{HTML}{F5F5F5}
\definecolor{faGrayLine}{HTML}{666666}
\definecolor{shadowCol}{HTML}{CCCCCC}

\title{Compiler-First State Space Duality and Portable $O(1)$ Autoregressive Caching}

\author{\name Cosmo Santoni \email cosmo.santoni@imperial.ac.uk \\
    \addr Imperial College London
    \AND
    \name Anmol Thapar \email a.thapar@imperial.ac.uk \\
    \addr Imperial College London
}

\begin{document}

\maketitle

\begin{abstract}
State-space models (SSMs) such as Mamba-2 are typically released with fused CUDA and Triton kernels, tying practical inference to NVIDIA CUDA devices or separate kernel ports. State space duality (SSD) exposes four compiler-facing properties: a diagonal state matrix, a chunkable recurrence, einsum-dominated compute, and static control flow. Those properties are sufficient for the fusion and tiling passes in XLA (Accelerated Linear Algebra) to produce competitive code without custom kernels. An SSD inference path expressed in standard JAX primitives, with the autoregressive cache registered as a JAX PyTree (a nested container that JAX traces as part of the compiled program), reaches roofline-consistent utilisation on both Google Cloud TPU v6e and NVIDIA L40S from a single source. On TPU v6e, single-stream prefill reaches approximately $140$~TFLOPS, or $15\%$ model FLOP utilisation (MFU), the batch-$1$ roofline ceiling, and cached decode reaches up to $64\%$ hardware bandwidth utilisation (HBU). The same source runs unmodified on NVIDIA L40S, where cached decode throughput is sequence-length independent across all five model scales ($130$M to $2.7$B). WikiText-103 perplexity matches the Triton reference within $\pm 0.0005$ points across all five model scales ($130$M to $2.7$B), and hidden states agree to float32 rounding tolerance. The implementation is available at \url{https://github.com/CosmoNaught/mamba2-jax}.
\end{abstract}

\section{Introduction}
\label{sec:intro}

State-space models (SSMs)~\citep{gu2023mamba, dao2024transformers, mamba3} offer a linear-time alternative to attention~\citep{vaswani2017attention} for long-sequence modelling. Reported throughput in the SSM literature depends on hand-written CUDA and Triton kernels integrated with the model code, and ports to targets such as AMD ROCm~\citep{mamba_amd} and Apple Metal Performance Shaders (MPS)~\citep{mamba_macos} are maintained as separate community forks with their own kernel codepaths. Here, deployment means running a pretrained SSM inference path on the accelerator available for serving or evaluation, with prompt tokens, weights, and the fixed SSM state resident on that device. The operational decision is whether the same checkpoint can be served or benchmarked on the available hardware without changing model semantics. Under the canonical kernelized implementations, users without the matching CUDA stack must rely on a separate port or rewrite kernels, because the model code bypasses the compiler stack that would otherwise supply hardware portability.

Mamba-2~\citep{dao2024transformers} restructures the recurrence into the state space duality (SSD) algorithm. SSD has four compiler-facing properties: (i) the state matrix $A$ is a diagonal scalar per head; (ii) the recurrence is decomposed into fixed-size chunks of $L$ tokens; (iii) the heavy computation is a small set of batched einsum contractions; and (iv) the control flow is static at compile time, including the lower-triangular causal mask. These four properties are the \emph{structural conditions} of SSD. They map onto fusion and tiling passes in XLA (Accelerated Linear Algebra)~\citep{abadi2016tensorflow}, so the algorithm can be expressed in standard primitives rather than in a hand-tuned kernel, provided the JAX expression preserves the conditions at the primitive level.

The JAX implementation includes chunked-parallel prefill and cached autoregressive decoding, registers the autoregressive cache as a JAX PyTree (a nested container whose array leaves participate in JAX tracing), and benchmarks the resulting binary on Google Cloud TPU v6e and NVIDIA L40S. It reaches roofline-consistent utilisation on both targets from a single source. WikiText-103 perplexity matches the Triton reference within $5 \times 10^{-4}$ across all five checkpoints.

\paragraph{Technical results.}
\begin{enumerate}[leftmargin=2em]
    \item SSD-to-XLA compatibility depends on four structural conditions, and the implementation preserves them through explicit shaping, static masking, and precision choices.
    \item The Mamba-2 implementation carries an $O(1)$ autoregressive cache through compiled on-device control flow as a registered JAX PyTree, supporting CPU, GPU, and TPU from a single source with no host synchronisation during generation.
    \item On TPU v6e, single-stream prefill reaches ${\sim}140$~TFLOPS, or $15\%$ model FLOP utilisation (MFU), and decode reaches $64\%$ hardware bandwidth utilisation (HBU) at the batch-$1$ roofline ceilings. On NVIDIA L40S, the same source reproduces sequence-length-independent cached decode across all five model scales.
    \item WikiText-103 validation perplexity matches the Triton reference (\texttt{mamba\_ssm} v$2.2.2$) within $\pm 0.0005$ points across all five model scales, and hidden states agree to float32 rounding tolerance.
\end{enumerate}

\section{Related Work}
\label{sec:related}

\paragraph{Kernelised SSM implementations.}
The reference Mamba implementation~\citep{gu2023mamba} ships fused CUDA and Triton~\citep{tillet2019triton} kernels, in the lineage of IO-aware fused attention~\citep{dao2022flashattention}, that are integral to its reported throughput. Mamba-2~\citep{dao2024transformers} extends this with Triton kernels tailored to the chunked structure of the SSD algorithm. Mamba-3~\citep{mamba3} is concurrent and was unavailable with open weights at the time of writing. Ports to AMD ROCm~\citep{mamba_amd} and Apple MPS~\citep{mamba_macos} are maintained as separate community forks with their own kernel codepaths. Linear attention recasts attention itself as a recurrence with $O(1)$ per-step decode~\citep{katharopoulos2020transformers}, and gated linear attention reaches high training throughput through hand-written chunk-parallel kernels~\citep{yang2024gated}; both exhibit the kernel dependency that the compiler-first treatment of SSD removes.

\paragraph{JAX implementations.}
Earlier JAX SSM work targets Mamba-1~\citep{mamba_minimal_jax, mamba_jax_vvvm} or provides minimal Mamba-2 forward passes without an autoregressive cache or performance evaluation~\citep{mamba2_minimal_jax}; the structured state-space family also has a JAX-native parallel-scan implementation in S5~\citep{smith2023s5}, which reached long-sequence state of the art on the Long Range Arena benchmark~\citep{tay2021lra}. Community reports indicate that pure JAX SSM implementations without custom kernels were too slow for practical training~\citep{jax_mamba_discussion}.

\paragraph{Relation to Bonsai.} The Mamba-2 module evaluated here is part of Bonsai~\citep{bonsai}, a library of minimal, dependency-light JAX model implementations, and uses its module structure and registered-PyTree cache; the evaluated Bonsai revisions are \texttt{a907b75} for core code and \texttt{d8f8d11} for caching. The additions over Bonsai are a structural-conditions analysis for XLA codegen, cross-hardware roofline characterisation on TPU v6e and L40S, implementation-choice ablations, and WikiText-103 perplexity and float32 numerical-parity validation against the Triton reference.

\paragraph{Compiler-first inference.}
JAX's design~\citep{jax2018github} composes standard primitives and delegates device-specific code generation to XLA. \citet{pope2023efficiently} demonstrate high-efficiency Transformer inference on TPUs via XLA-compiled code with no custom kernels. The earlier structured state space literature~\citep{gu2022efficiently} relied on custom CUDA kernels for the selective scan, and the algebraic structure of SSD removes that dependency. PyTorch's torch.compile path (Dynamo and Inductor)~\citep{ansel2024pytorch2} offers a comparable compilation route on CUDA targets, but has no mature TPU backend at the time of writing.

\section{Method}
\label{sec:method}

SSD is compiler-friendly only when its algebraic constraints survive the JAX front end. The JAX path enforces three requirements. The recurrence remains diagonal, chunked, einsum-dominated, and statically masked; the primitive graph exposes those properties to XLA; and the autoregressive state is part of the compiled loop rather than Python-side metadata. The recurrence notation follows \citet{dao2024transformers}.

\subsection{State-space duality}
\label{sec:ssd_background}

A continuous-time SSM maps an input signal $x(t) \in \mathbb{R}^N$ to an output $y(t)$ through a latent state $h(t) \in \mathbb{R}^N$:
\begin{equation}
h'(t) = A h(t) + B x(t), \qquad y(t) = C h(t) + D x(t),
\end{equation}
with $A \in \mathbb{R}^{N \times N}$, $B \in \mathbb{R}^{N \times 1}$, $C \in \mathbb{R}^{1 \times N}$, $D \in \mathbb{R}$. Discretisation by zero-order hold with step size $\Delta$ yields the discrete recurrence
\begin{equation}
h_t = \bar{A} h_{t-1} + \bar{B} x_t, \qquad y_t = C h_t + D x_t.
\end{equation}
Mamba-2~\citep{dao2024transformers} makes $B$, $C$, and $\Delta$ input-dependent and restricts $A$ to a diagonal scalar per head. Unrolled over a chunk of $L$ tokens, the discrete recurrence admits the dual form
\begin{equation}
Y_{\text{diag}} = (\mathcal{L} \odot C B^\top) X,
\label{eq:ssd}
\end{equation}
where $\mathcal{L}$ is a lower-triangular matrix of accumulated decay factors obtained by exponentiating a segment-wise prefix sum of $A \cdot \Delta$ over each chunk. Inter-chunk state propagation is a separate sequential recurrence over chunk-level summary states whose FLOP count is small relative to the intra-chunk matmuls.

\subsection{Structural conditions of SSD}
\label{sec:conditions}

SSD has four compiler-facing properties.

\textbf{Diagonal state matrix.} Restricting $A$ to a diagonal scalar per head reduces the matrix exponential in the discretised recurrence to a scalar exponential, so that the cumulative decay over a chunk is the exponential of a segment-wise prefix sum. The recurrence unrolls analytically across a fixed window.

\textbf{Chunked recurrence.} The full sequence is partitioned into fixed-size chunks of $L$ tokens ($L = 256$ throughout, the default of \citet{dao2024transformers}). Within each chunk the sequential recurrence unrolls into the parallel matrix computation of Eq.~\ref{eq:ssd}; across chunks, a lightweight scan propagates summary state. Chunk size controls the balance between arithmetic intensity and sequential overhead. Larger $L$ raises the arithmetic intensity of intra-chunk matmuls and the working set, while smaller $L$ shifts the balance toward inter-chunk sequential overhead.

\textbf{Einsum-dominated compute.} The intra-chunk output is a single batched einsum over axes (batch, chunk, sequence-within-chunk, head, state). The accompanying element-wise operations (softplus, exponentials, lower-triangular masking) are memory-bound chains that compose with the einsum into a single fused region of the compute graph.

\textbf{Static control flow.} The chunk count, kernel size, and head structure are known at compile time, and the lower-triangular causal mask is a constant of $L$ rather than a function of token content. The recurrence can therefore be expressed without data-dependent control flow at the front end.

\subsection{Mapping SSD onto JAX primitives}
\label{sec:mapping}

The structural conditions are necessary but not sufficient. A JAX expression that hides them behind dynamic loops, irregular indexing, or low-precision state updates will not compile to the fused XLA graph needed for high throughput. The implementation uses four primitive-level choices. Ablations quantify the cost of reverting masking, loop placement, and decay precision. The SSD core function is approximately sixty lines of Python~\citep{mamba2jax}; its contractions are batched over batch, chunk, sequence, head, and state axes.

\textbf{Einsum shaping.} All heavy computation is expressed as batched einsum contractions. Inputs are reshaped so that batch, head, chunk, and sequence-within-chunk dimensions produce large contiguous matrix operands. The contractions then map directly onto tiled GEMM calls on the target's matrix units, and the surrounding element-wise chains of softplus, clip, and exp fuse into the same region.

\textbf{Static masking.} Applying the lower-triangular causal mask to a precomputed matrix gives XLA a static constant it can fold into the surrounding fusion chain of prefix sum, subtraction, mask, and exponentiation. Applying the same mask row by row inside a runtime loop produces bitwise-identical output but pays an $82.8\%$ throughput penalty, attributable to the fusion graph breaking at the loop boundary (Table~\ref{tab:ablation_masking}).

\textbf{Compiled on-device loops.} Autoregressive decoding is wrapped in a compiled on-device loop, so that the loop body, the cache update, and the deterministic on-device argmax execute as one compiled XLA program. Driving the same loop from Python and synchronising on each iteration costs $2.4\times$ throughput at the $130$M scale. The gap arises from host--device round-trips and dissolves above $780$M parameters where per-step compute dominates the round-trip cost (Table~\ref{tab:host_vs_scan}).

\textbf{Precision management.} Four precision rules are required for downstream parity. Residual connections are kept in float32 to prevent accumulation drift through the layer stack. Decay parameters are held in log-space float32 and exponentiated at compute time to avoid underflow. Normalisation layers cast inputs to float32 for the variance reduction and cast back. The default matmul precision is set to the highest available mode for correctness validation, suppressing TensorFloat-32 (TF32)-style hardware rounding on NVIDIA targets. Using bfloat16 for decay exponentiation alone produces a maximum absolute logit error of $0.013$ at $130$M parameters, large enough to shift the output distribution (Table~\ref{tab:ablation_precision}).

\subsection{The $O(1)$ state cache}
\label{sec:cache}

SSMs maintain a fixed-size hidden state $h \in \mathbb{R}^{H \times P \times N}$ that summarises the entire prefix, so generating the next token requires a depthwise convolution update over a sliding window of $k{-}1$ cached inputs and a single recurrence step $h_t = \bar{A} h_{t-1} + \bar{B} x_t$. Both operations are $O(1)$ in the prefix length.

The per-layer SSM and convolution states are stored in one dataclass registered as a JAX PyTree, a nested container whose array leaves participate in JAX tracing. Just-in-time (JIT) compilation and on-device control flow then trace the cache into the compiled loop without host round-trips. Prefill initialises the cache with the final chunk state for each layer, and the per-token decode body updates the convolution window, applies one SSM step, and emits the next token on device. The host--device boundary is one compiled XLA launch: the Python host is inactive during generation, while the cache, parameters, and argmax remain on the accelerator.

\begin{algorithm}[htbp]
\DontPrintSemicolon
\SetAlgoLined
\KwIn{Token ids $x_{1:T}$, model parameters $\theta$, chunk size $L$}
\KwOut{Logits $\hat{y}_{1:T}$, per-layer final chunk states $\{h^{(\ell)}\}$}
Embed and reshape: $X \gets \text{Embed}(x) \in \mathbb{R}^{B \times N_c \times L \times D}$ \tcp*{$N_c = T/L$ chunks}
\For{layer $\ell = 1, \ldots, N_{\text{layers}}$}{
    Project: $B_t, C_t, \Delta, X_{\text{proj}} \gets \text{InputProj}^{(\ell)}(X)$\;
    Discretise: $\bar{A} \gets \exp(\texttt{softplus}(A^{(\ell)}_{\log}) \cdot \Delta)$ \tcp*{float32 upcast}
    Decay matrix: $\mathcal{L} \gets \texttt{tril}(\exp(\texttt{segsum}(\log \bar{A})))$\;
    Intra-chunk: $Y_{\text{diag}} \gets (\mathcal{L} \odot C_t B_t^\top)\, X_{\text{proj}}$ \tcp*{batched einsum}
    State accumulate: $S \gets \texttt{einsum}(B_t, \bar{A}, X_{\text{proj}}) \to (B, N_c, H, P, N)$\;
    Inter-chunk scan: $S' \gets \texttt{einsum}(\bar{A}_{\text{chunk}}, S)$ \tcp*{sequential over chunks}
    Combine: $Y \gets Y_{\text{diag}} + Y_{\text{cross}}(S', C_t)$\;
    Store $h^{(\ell)} \gets S'[:, -1] \in \mathbb{R}^{B \times H \times P \times N}$ \tcp*{final chunk state $\to$ cache}
}
$\hat{y}_{1:T} \gets \text{LMHead}(Y)$\;
\caption{SSD prefill (one call, chunked parallel)}
\label{alg:prefill}
\end{algorithm}

\begin{algorithm}[htbp]
\DontPrintSemicolon
\SetAlgoLined
\KwIn{Prompt ids $x_{1:P}$, generation length $G$, model parameters $\theta$}
\KwOut{Generated token ids $\hat{x}_{P+1:P+G}$}
$\hat{y}_{1:P},\; \mathcal{C} \gets \textsc{Prefill}(x_{1:P}, \theta)$ \tcp*{Algorithm~\ref{alg:prefill}; init cache}
$\hat{x}_{P+1} \gets \arg\max\, \hat{y}_P$\;
\For(\tcp*[f]{compiled \texttt{fori\_loop}, on-device}){$t = P{+}1, \ldots, P{+}G{-}1$}{
    $u_t \gets \text{Embed}(\hat{x}_t)$\;
    \For{layer $\ell = 1, \ldots, N_{\text{layers}}$}{
        $u_t^{(\ell)} \gets \text{InProj}^{(\ell)}(u_t)$ \tcp*{project to inner dim}
        Update conv state: $\mathcal{C}.\text{conv}^{(\ell)} \gets \texttt{roll\_and\_insert}(u_t^{(\ell)})$\;
        $z_t \gets \text{DepthwiseConv}(\mathcal{C}.\text{conv}^{(\ell)})$\;
        Project: $B_t, C_t, \Delta_t \gets \text{SSMProj}^{(\ell)}(z_t)$\;
        Discretise: $\bar{A} \gets \exp(\texttt{softplus}(A^{(\ell)}_{\log}) \cdot \Delta_t),\; \bar{B} \gets \Delta_t \cdot B_t$\;
        SSM step: $\mathcal{C}.\text{ssm}^{(\ell)} \gets \bar{A}\, \mathcal{C}.\text{ssm}^{(\ell)} + \bar{B}\, z_t$\;
    }
$\hat{x}_{t+1} \gets \arg\max\, \text{LMHead}(y_t)$ \tcp*{on-device}
}
\caption{Cached autoregressive decode ($O(1)$ per step)}
\label{alg:decode}
\end{algorithm}

\begin{figure}[htbp]
    \centering
    \begin{tikzpicture}[
    >=Stealth,
    font=\sffamily,
    tensor/.style={
        rectangle, draw=faGreenLine, fill=faGreen,
        rounded corners=14pt, ultra thick,
        minimum width=3.6cm, minimum height=1.0cm,
        align=center, font=\sffamily\large\bfseries
    },
    shadow/.style={
        rectangle, fill=shadowCol, rounded corners=4pt,
        minimum width=5.8cm, minimum height=1.2cm
    },
    opBlue/.style={
        rectangle, draw=faBlueLine, fill=faBlue,
        rounded corners=4pt, ultra thick,
        minimum width=5.8cm, minimum height=1.2cm, align=center
    },
    opYellow/.style={
        rectangle, draw=faYellowLine, fill=faYellow,
        rounded corners=4pt, ultra thick,
        minimum width=5.8cm, minimum height=1.4cm, align=center
    },
    opRed/.style={
        rectangle, draw=faRedLine, fill=faRed,
        rounded corners=4pt, ultra thick,
        minimum width=5.8cm, minimum height=1.2cm, align=center
    },
    opPurple/.style={
        rectangle, draw=faPurpleLine, fill=faPurple,
        rounded corners=4pt, ultra thick,
        minimum width=5.8cm, minimum height=1.2cm, align=center
    },
    flow/.style={->, ultra thick, draw=black!60,
        shorten >=2pt, shorten <=2pt},
    dataflow/.style={<->, thick, draw=faGrayLine, densely dashed,
        shorten >=3pt, shorten <=3pt},
    dimlabel/.style={font=\sffamily\scriptsize, color=faGrayLine}
]


\node[tensor]  (xt)   at (0,  0.0)  {$\hat{x}_t$};
\node[opBlue]  (proj) at (0, -2.2)  {
    \textbf{Input Projection}\\[3pt]
    \small$u_t \gets \mathrm{InProj}(\hat{x}_t)$};
\node[opYellow](conv) at (0, -4.6)  {
    \textbf{Depthwise Conv}\\[3pt]
    \small$\mathcal{C}.\mathrm{conv}^{(\ell)} \gets \mathrm{shift}(\mathcal{C}.\mathrm{conv}^{(\ell)},\, u_t)$\\[1pt]
    \small$z_t \gets \mathrm{DepthwiseConv}(\mathcal{C}.\mathrm{conv}^{(\ell)})$};
\node[opRed]   (ssm)  at (0, -7.6)  {
    \textbf{SSM Step}\\[3pt]
    \small$\mathcal{C}.\mathrm{ssm}^{(\ell)} \gets \bar{A}\, \mathcal{C}.\mathrm{ssm}^{(\ell)} + \bar{B}z_t$};
\node[opPurple](head) at (0,-10.0)  {
    \textbf{LM Head \& Argmax}\\[3pt]
    \small$\hat{x}_{t+1} \gets \arg\max(\mathrm{LMHead}(y_t))$};
\node[tensor]  (xt1)  at (0,-12.2)  {$\hat{x}_{t+1}$};

\begin{scope}[on background layer]
    \foreach \n in {proj,conv,ssm,head}{
        \node[shadow] at ([xshift=3pt,yshift=-3pt]\n) {};
    }
\end{scope}

\draw[flow] (xt)   -- (proj);
\draw[flow] (proj) -- node[right=4pt, fill=white, rounded corners=2pt,
    font=\small\sffamily, inner sep=2pt] {$u_t$} (conv);
\draw[flow] (conv) -- node[right=4pt, fill=white, rounded corners=2pt,
    font=\small\sffamily, inner sep=2pt] {$z_t$} (ssm);
\draw[flow] (ssm)  -- node[right=4pt, fill=white, rounded corners=2pt,
    font=\small\sffamily, inner sep=2pt] {$y_t$} (head);
\draw[flow] (head) -- (xt1);


\begin{scope}[shift={(5.8, -4.6)}]
    \fill[shadowCol, rounded corners=2pt]
        (-0.8+3pt,-0.9-3pt) rectangle (0.8+3pt, 0.9-3pt);
    \fill[faYellow] (-0.8,-0.9) rectangle (0.8, 0.9);
    
    \foreach \x in {-0.4, 0.0, 0.4}{
        \draw[faYellowLine, thin] (\x,-0.9) -- (\x, 0.9);}
    \foreach \y in {-0.3, 0.3}{
        \draw[faYellowLine, thin] (-0.8,\y) -- (0.8,\y);}
    \draw[faYellowLine, thick] (-0.8,-0.9) rectangle (0.8, 0.9);

    \node[above=4pt, font=\sffamily\small\bfseries\color{faYellowLine!80!black}]
        (conv_state_lbl) at (0, 0.9) {Conv State};

    \draw[faYellowLine]
        (-0.8,-1.10) -- (-0.8,-1.00)
        (-0.8,-1.10) -- ( 0.8,-1.10)
        ( 0.8,-1.10) -- ( 0.8,-1.00);
    \node[below=1pt, dimlabel] at (0,-1.10)
        {$D_\text{conv}{\times}(k{-}1)$};

    \draw[faYellowLine]
        (0.95,-0.9) -- (0.85,-0.9)
        (0.95,-0.9) -- (0.95, 0.9)
        (0.95, 0.9) -- (0.85, 0.9);
    \node[right=1pt, dimlabel] at (0.95, 0) {$B$};

    \coordinate (csnw) at (-1.55,  1.50);
    \coordinate (csse) at ( 2.40, -1.55);
\end{scope}

\begin{scope}[shift={(5.8, -7.6)}]
    \fill[shadowCol, rounded corners=2pt]
        (-1.6+3pt,-0.9-3pt) rectangle (1.6+3pt, 0.9-3pt);
    \fill[faRed] (-1.6,-0.9) rectangle (1.6, 0.9);
    
    \foreach \x in {-1.06, -0.53, 0.0, 0.53, 1.06}{
        \draw[faRedLine, thin] (\x,-0.9) -- (\x, 0.9);}
    \foreach \y in {-0.3, 0.3}{
        \draw[faRedLine, thin] (-1.6,\y) -- (1.6,\y);}
    \draw[faRedLine, thick] (-1.6,-0.9) rectangle (1.6, 0.9);

    \node[draw=faRedLine, fill=faRed!60!white, rounded corners=0pt,
          font=\sffamily\scriptsize\bfseries\color{faRedLine!80!black},
          inner sep=3pt] at (0, 0) {$O(1)$ fixed};

    \draw[faRedLine]
        (-1.6,-1.10) -- (-1.6,-1.00)
        (-1.6,-1.10) -- ( 1.6,-1.10)
        ( 1.6,-1.10) -- ( 1.6,-1.00);
    \node[below=1pt, dimlabel] (ssm_dim_lbl) at (0,-1.10)
        {$H{\times}P{\times}N$};

    \node[below=6pt, font=\sffamily\small\bfseries\color{faRedLine!80!black}]
        (ssm_state_lbl) at (ssm_dim_lbl.south) {SSM State};

    \draw[faRedLine]
        (1.75,-0.9) -- (1.65,-0.9)
        (1.75,-0.9) -- (1.75, 0.9)
        (1.75, 0.9) -- (1.65, 0.9);
    \node[right=1pt, dimlabel] at (1.75, 0) {$B$};

    \coordinate (ssnw) at (-1.80,  1.10);
    \coordinate (ssse) at ( 2.40, -2.10);
\end{scope}

\draw[dataflow] (conv.east) -- (4.8, -4.6)
    node[midway, above=2pt, font=\small\sffamily\color{faGrayLine}]
    {\textbf{R/W}};
\draw[dataflow] (ssm.east) -- (4.0, -7.6)
    node[midway, above=2pt, font=\small\sffamily\color{faGrayLine}]
    {\textbf{R/W}};


\draw[flow, blue!60!black, line width=2.2pt, rounded corners=10pt]
    (xt1.west) -- ++(-2.8, 0) -- ++(0, 12.2) -- (xt.west);

\node[fill=blue!10, draw=blue!50!black, rounded corners=8pt,
      font=\sffamily\small\bfseries\color{blue!70!black},
      inner xsep=8pt, inner ysep=4pt, rotate=90]
    at (-3.75, -6.1)
    {Compiled on-device loop \enspace
     \normalfont\scriptsize$(t = P{+}1,\,\dots,\,P{+}G{-}1)$};


\begin{scope}[on background layer]

    \node[rectangle, draw=orange!55!black, dashed, thick,
          fill=orange!5, rounded corners=8pt,
          fit=(csnw)(csse)(ssnw)(ssse)(conv_state_lbl)(ssm_state_lbl),
          inner xsep=14pt, inner ysep=10pt] (cache) {};

    \node[fill=orange!15, draw=orange!60!black, rounded corners=6pt,
          font=\sffamily\small\bfseries\color{orange!80!black},
          inner xsep=8pt, inner ysep=3pt]
        at (cache.north) {SSM cache, $O(1)$ JAX PyTree};

    \coordinate (left_margin) at (-4.2, -6.1);

    \node[rectangle, draw=teal!60!black, dashed, very thick,
          fill=teal!3, rounded corners=10pt,
          fit=(xt)(xt1)(cache)(left_margin),
          inner xsep=16pt, inner ysep=20pt] (device) {};

    \node[anchor=north west,
          font=\sffamily\normalsize\bfseries\color{teal!65!black},
          inner sep=10pt] at (device.north west)
        {Compiled XLA Device Execution \enspace
         \normalfont\small\color{teal!55!black}(TPU / GPU)};

    \node[rectangle, draw=faGrayLine, fill=faGray, rounded corners=5pt,
          thick, minimum width=5.8cm, minimum height=1.0cm, align=center]
        (host) at ([yshift=1.8cm]device.north -| xt.north)
        {\textbf{Python Host Context}\\[2pt]
         \small\color{faGrayLine} Suspended during decode};

    \draw[->, thick, draw=faGrayLine]
        (host.south) -- (host.south |- device.north)
        node[midway, right=3pt, font=\small\sffamily] {Launch HLO};

\end{scope}

\end{tikzpicture}
    \caption{Host--device boundary for compiled autoregressive decoding. The SSM cache is a JAX PyTree, so the compiled loop updates the $O(1)$ SSM and convolution state on device. The Python host launches the compiled XLA program once and remains inactive until generation completes.}
    \label{fig:architecture}
\end{figure}

\section{Evaluation}
\label{sec:evaluation}

\subsection{Setup}
\label{sec:setup}

\textbf{Hardware.} Primary benchmarks run on a single Google Cloud TPU v6e chip~\citep{tpu_v6e_specs} with $918$ TFLOPS bfloat16 (BF16) peak and $1600$ GB/s HBM bandwidth. GPU benchmarks run on a single NVIDIA L40S~\citep{l40s_specs} with $362$ TFLOPS BF16 peak and $864$ GB/s GDDR6 bandwidth. The same source code runs on both platforms.

\textbf{Models.} Five pretrained Mamba-2 checkpoints from HuggingFace, spanning $130$M to $2.7$B parameters, all with state size $128$, head dimension $64$, expand factor $2$, and conv kernel size $4$. Chunk size is fixed at $L = 256$. The checkpoints use the original \texttt{state-spaces/mamba2-*} weights.

\textbf{Inference protocol.} Throughput, memory, and utilisation experiments use single-stream decoding (batch size $1$). Each decode step is a forward pass followed by a deterministic on-device argmax inside the compiled decode loop, with no sampling applied. Throughput is generated tokens per wall-clock second, measured with a host wall-clock timer around calls forced to complete by an explicit synchronisation barrier. All timings are averaged over five runs after JIT warm-up, with standard deviations below $0.3\%$ of the mean.

\textbf{Training-step protocol.} The reduced training-step comparison against the Triton reference covers the three smallest checkpoints ($130$M--$780$M) at batch size $1$ and sequence lengths $\{512, 1024, 2048\}$; the $1.3$B and $2.7$B checkpoints and longer sequences exceed L40S memory under the JAX path and are omitted. Each timed step is a forward $+$ backward pass, both implementations load the same HuggingFace checkpoints, and cells are the mean of ten timed steps after ten warm-ups.

\textbf{Metrics.} Model FLOP utilisation (MFU) and hardware bandwidth utilisation (HBU) follow \citet{chowdhery2023palm},
\begin{align}
\text{MFU} &= \frac{F_{\text{XLA}} \;/\; t_{\text{wall}}}{\text{peak TFLOPS} \times 10^{12}}, \label{eq:mfu} \\
\text{HBU} &= \frac{B_{\text{XLA}} \;/\; t_{\text{wall}}}{\text{peak BW} \times 10^{9}}, \label{eq:hbu}
\end{align}
where $F_{\text{XLA}}$ and $B_{\text{XLA}}$ are the FLOP and byte-access counts reported by the XLA cost analysis for the lowered XLA program. $F_{\text{XLA}}$ is exact for einsum-dominated workloads. $B_{\text{XLA}}$ is an unfused byte count, so fusion may collapse intermediate buffers, and reported HBU is therefore an upper bound on true bandwidth efficiency.

\textbf{Baseline.} The non-cached baseline recomputes the full forward pass over the entire token sequence at every decode step. It uses the same model functions on the same hardware, with the SSM cache disabled.

\textbf{Software versions.} Runs use JAX 0.9.0, jaxlib 0.9.0.1, Python 3.12, Flax NNX 0.12.4, PyTorch 2.10.0, and \texttt{mamba\_ssm} v$2.2.2$; the project repository contains the benchmark scripts, configuration files, and invocation commands~\citep{mamba2jax}.

\subsection{Single-Stream Autoregressive Throughput (TPU v6e)}
\label{sec:tpu_throughput}

\begin{figure}[htbp]
\centering
\includegraphics[width=\textwidth]{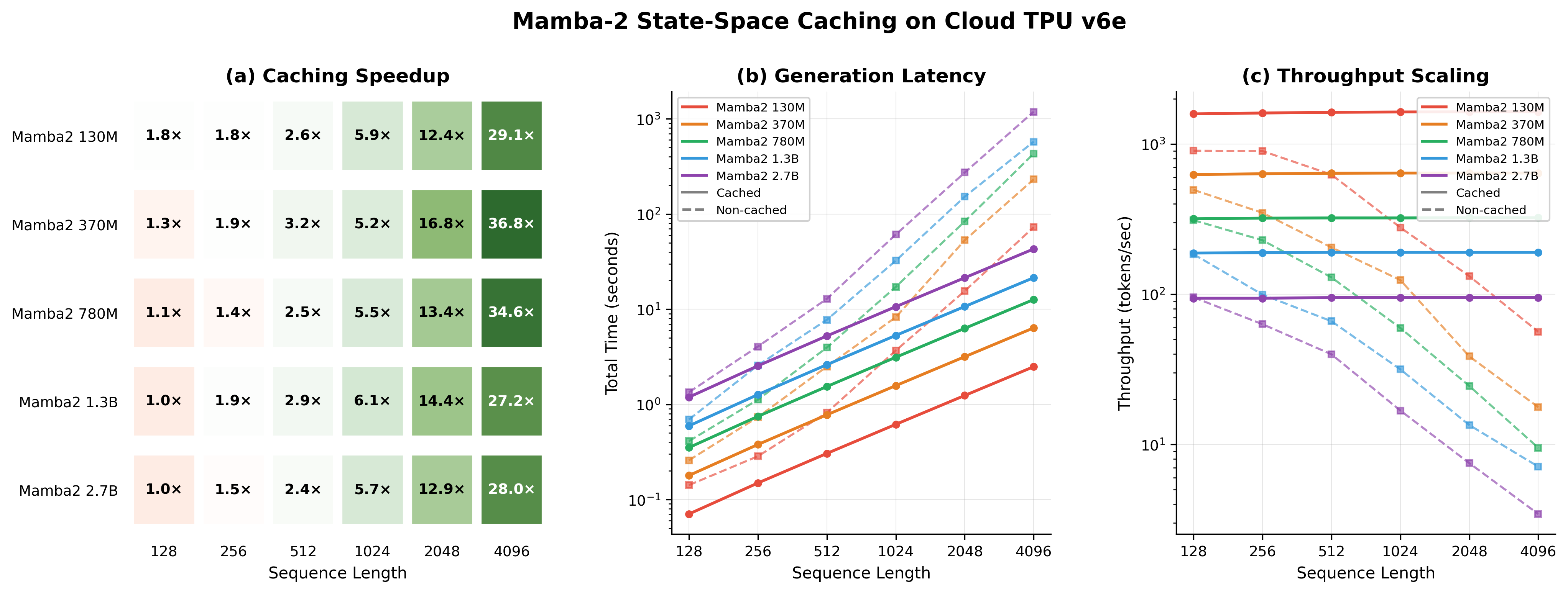}
\caption{Autoregressive generation on Cloud TPU v6e across five model scales and six sequence lengths. \textbf{(a)}~Speedup from caching. \textbf{(b)}~Generation latency. Cached (solid) grows linearly, non-cached (dashed) grows quadratically. \textbf{(c)}~Per-step throughput. Cached throughput is flat regardless of sequence length, and non-cached throughput collapses.}
\label{fig:scaling}
\end{figure}

Caching produces sequence-length-independent per-step throughput at every model scale (Figure~\ref{fig:scaling}). The compiled on-device loop outperforms the host-driven cached loop by $2.4\times$ at the $130$M scale ($1588$ versus $662$ tokens/s). Above $780$M parameters the per-step compute dominates the round-trip overhead and the two paths converge (Table~\ref{tab:host_vs_scan}).

\begin{table}[htbp]
\centering
\begin{tabular}{llccc}
\toprule
& & \multicolumn{3}{c}{\textbf{Throughput (Tokens/Second)}} \\
\cmidrule(lr){3-5}
\textbf{Model} & \textbf{Method} & \textbf{128} & \textbf{1024} & \textbf{4096} \\
\midrule
130M & Cached (scan)  & 1588 & 1635 & 1641 \\
     & Cached (host)  & 662  & 729  & 751 \\
     & Non-Cached     & 903  & 278  & 56 \\
\midrule
370M & Cached (scan)  & 626  & 641  & 641 \\
     & Cached (host)  & 392  & 391  & 390 \\
     & Non-Cached     & 495  & 124  & 18 \\
\midrule
780M & Cached (scan)  & 318  & 322  & 323 \\
     & Cached (host)  & 325  & 326  & 327 \\
     & Non-Cached     & 311  & 60   & 9 \\
\midrule
1.3B & Cached (scan)  & 188  & 190  & 190 \\
     & Cached (host)  & 192  & 192  & 192 \\
     & Non-Cached     & 185  & 32   & 7 \\
\midrule
2.7B & Cached (scan)  & 94   & 95   & 95 \\
     & Cached (host)  & 97   & 96   & 96 \\
     & Non-Cached     & 95   & 17   & 3 \\
\bottomrule
\end{tabular}
\caption{Decode strategy comparison on TPU v6e (batch size 1). Small inversions
between cached scan and cached host at 780M (318 vs 325) and 2.7B (94 vs 97) are
within measurement variance; both models are per-step-compute dominated, so the two
paths have converged.}
\label{tab:host_vs_scan}
\end{table}

\subsection{Peak Memory (TPU v6e)}

\begin{figure}[htbp]
\centering
\includegraphics[width=0.75\textwidth]{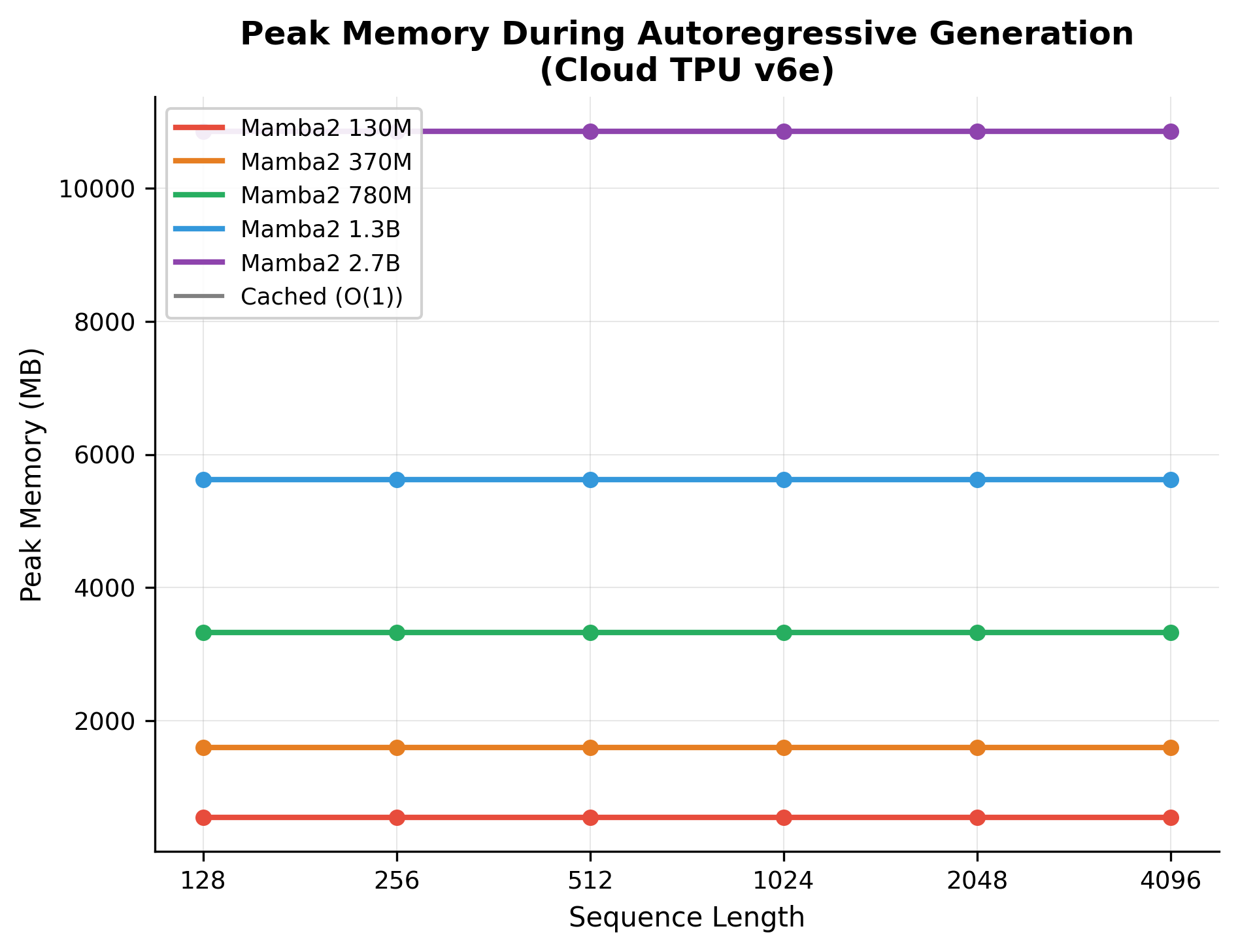}
\caption{Peak memory during autoregressive generation on Cloud TPU v6e. Cached
decoding (solid) is constant; non-cached (dashed) grows linearly. At sequence length
4096, the 2.7B non-cached path consumes over 16\,GB versus a constant 10.9\,GB
cached.}
\label{fig:memory}
\end{figure}

Cached decoding holds peak memory constant while the non-cached path grows linearly with sequence length (Figure~\ref{fig:memory}). Peak memory is measured via the JAX device runtime (peak-bytes-in-use counter),
reporting on-device HBM allocation. The cache stores per-layer SSM states of shape
$(B, H, P, N)$ and convolution states of shape $(B, D_{\text{conv}}, k{-}1)$, neither
of which depends on sequence length. The non-cached path materialises the full $(1,
L)$ token buffer and intermediate activations at each step, scaling linearly.

\subsection{Hardware Utilisation (TPU v6e, Single-Stream)}
\label{sec:utilisation}
\begin{figure}[htbp]
\centering
\includegraphics[width=\textwidth]{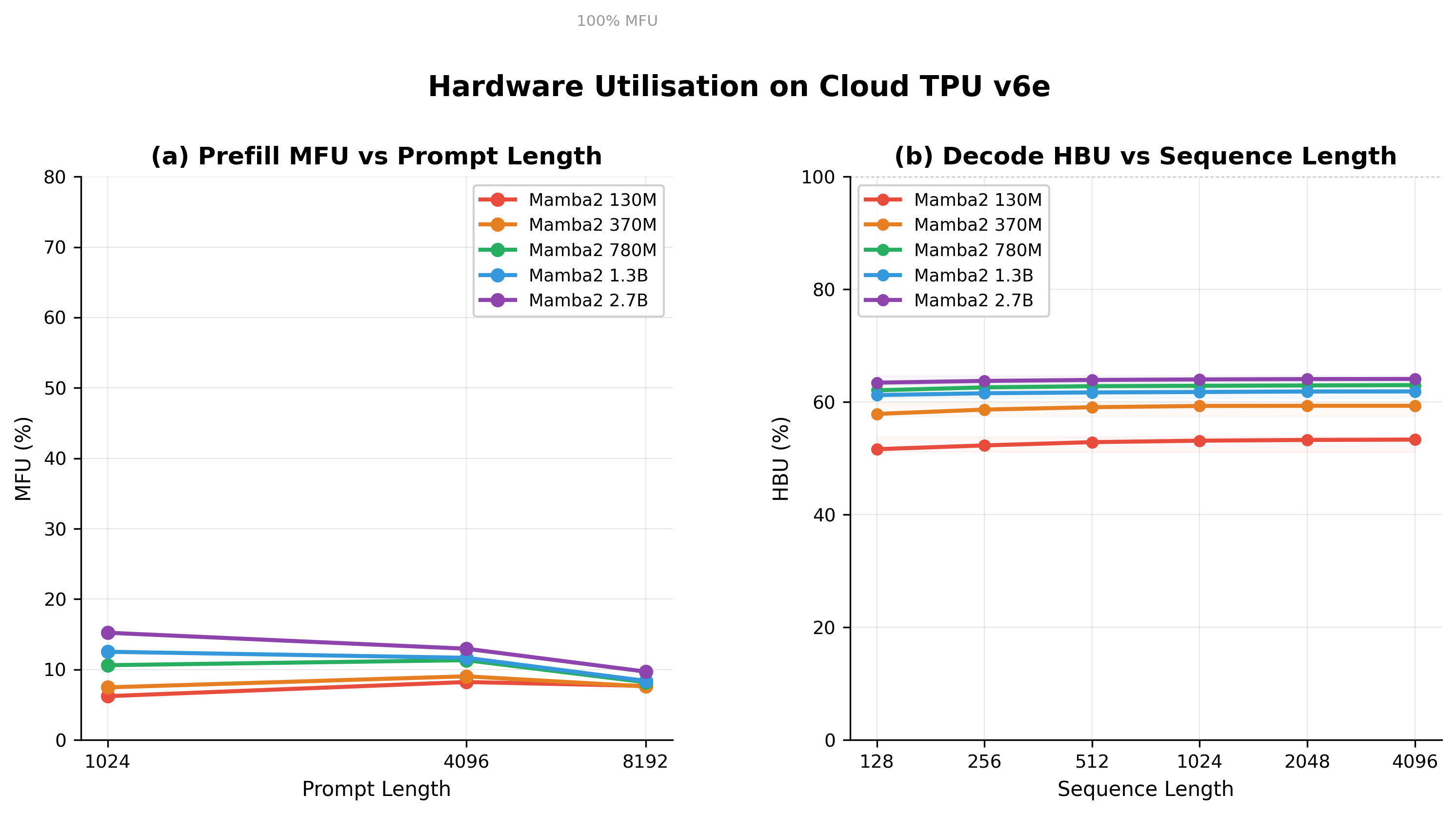}
\caption{Hardware utilisation on Cloud TPU v6e (batch size 1). \textbf{(a)}~Prefill
MFU versus model size at three prompt lengths. \textbf{(b)}~Decode HBU versus
sequence length. HBU varies by less than 1.7 percentage points across sequence
lengths for every model.}
\label{fig:utilisation}
\end{figure}

Single-sequence prefill is compute-bound and cached decode is memory-bandwidth bound (Figure~\ref{fig:utilisation}). Saturating the v6e's compute requires approximately $574$ FLOPs per byte. Batch size $1$ does not cross this threshold, so the observed $15\%$ prefill MFU matches the roofline ceiling~\citep{williams2009roofline} for this regime rather than an avoidable compiler gap. The MFU rise from $1024$ to $4096$ tokens at smaller model scales reflects better intra-chunk matmul tiling as the chunk count $N_c$ grows (Table~\ref{tab:prefill_mfu}). Beyond $4096$ tokens the sequential inter-chunk scan adds $O(N_c)$ serial dispatch overhead that reduces measured MFU at every model size.

\begin{table}[htbp]
\centering
\begin{tabular}{lccc}
\toprule
& \multicolumn{3}{c}{\textbf{Prefill MFU (\%) by Prompt Length}} \\
\cmidrule(lr){2-4}
\textbf{Model} & \textbf{1024} & \textbf{4096} & \textbf{8192} \\
\midrule
130M & 6.22 & 8.23 & 7.68 \\
370M & 7.47 & 9.04 & 7.60 \\
780M & 10.62 & 11.33 & 8.20 \\
1.3B & 12.53 & 11.67 & 8.39 \\
2.7B & 15.23 & 12.96 & 9.71 \\
\bottomrule
\end{tabular}
\caption{Prefill compute efficiency on TPU v6e (batch size 1, peak = 918 TFLOPS
BF16).}
\label{tab:prefill_mfu}
\end{table}

\begin{table}[htbp]
\centering
\begin{tabular}{lcccccc}
\toprule
& \multicolumn{6}{c}{\textbf{Decode HBU (\%) by Sequence Length}} \\
\cmidrule(lr){2-7}
\textbf{Model} & \textbf{128} & \textbf{256} & \textbf{512} & \textbf{1024} &
\textbf{2048} & \textbf{4096} \\
\midrule
130M & 51.62 & 52.29 & 52.87 & 53.13 & 53.26 & 53.32 \\
370M & 57.88 & 58.65 & 59.07 & 59.29 & 59.32 & 59.32 \\
780M & 62.07 & 62.59 & 62.80 & 62.87 & 62.93 & 62.99 \\
1.3B & 61.22 & 61.55 & 61.69 & 61.77 & 61.86 & 61.87 \\
2.7B & 63.43 & 63.74 & 63.91 & 64.00 & 64.06 & 64.08 \\
\bottomrule
\end{tabular}
\caption{Decode memory-bandwidth efficiency on TPU v6e (batch size 1, peak = 1600
GB/s).}
\label{tab:decode_hbu}
\end{table}

Decode HBU varies by less than $1.7$ percentage points across all sequence lengths for every model (Table~\ref{tab:decode_hbu}). Constant HBU across sequence lengths follows from the fixed-size cache: each decode step reads and writes the same SSM and convolution state regardless of prefix length.

\FloatBarrier
\subsection{Single-Stream Throughput on NVIDIA L40S}
\label{sec:l40}

\begin{table}[htbp]
\centering
\begin{tabular}{llccc}
\toprule
& & \multicolumn{3}{c}{\textbf{Throughput (Tokens/Second)}} \\
\cmidrule(lr){3-5}
\textbf{Model} & \textbf{Method} & \textbf{128} & \textbf{1024} & \textbf{4096} \\
\midrule
130M & Cached (scan)   & 240.2 & 267.1 & 314.2 \\
     & Cached (host)   & 178.4 & 141.9 & 188.5 \\
     & Non-Cached      & 203.3 & 115.8 & 20.3 \\
\midrule
370M & Cached (scan)   & 154.3 & 165.1 & 148.0 \\
     & Cached (host)   & 104.1 & 98.8  & 112.3 \\
     & Non-Cached      & 125.4 & 36.9  & 7.2 \\
\midrule
780M & Cached (scan)   & 110.2 & 106.4 & 108.0 \\
     & Cached (host)   & 107.2 & 118.5 & 99.6 \\
     & Non-Cached      & 97.3  & 20.4  & 3.9 \\
\midrule
1.3B & Cached (scan)   & 67.2  & 71.3  & 71.0 \\
     & Cached (host)   & 71.1  & 72.4  & 72.5 \\
     & Non-Cached      & 65.2  & 12.7  & 2.7 \\
\midrule
2.7B & Cached (scan)   & 35.4  & 36.3  & 36.1 \\
     & Cached (host)   & 37.2  & 37.1  & 37.1 \\
     & Non-Cached      & 34.8  & 6.7   & 1.5 \\
\bottomrule
\end{tabular}
\caption{Single-stream autoregressive decode throughput on NVIDIA L40S (batch size $1$). Sequence lengths mirror the TPU evaluation in Table~\ref{tab:host_vs_scan}.}
\label{tab:l40_throughput}
\end{table}

On L40S, cached decode also remains sequence-length independent (Table~\ref{tab:l40_throughput}). The host-driven loop incurs a measurable round-trip penalty at smaller model sizes and converges with the compiled loop at large scale, where per-step compute dominates. Absolute throughput follows the L40S's lower compute and bandwidth ceilings.

\subsection{Downstream Perplexity}
\label{sec:perplexity}

\begin{figure}[htbp]
\centering
\includegraphics[width=0.75\textwidth]{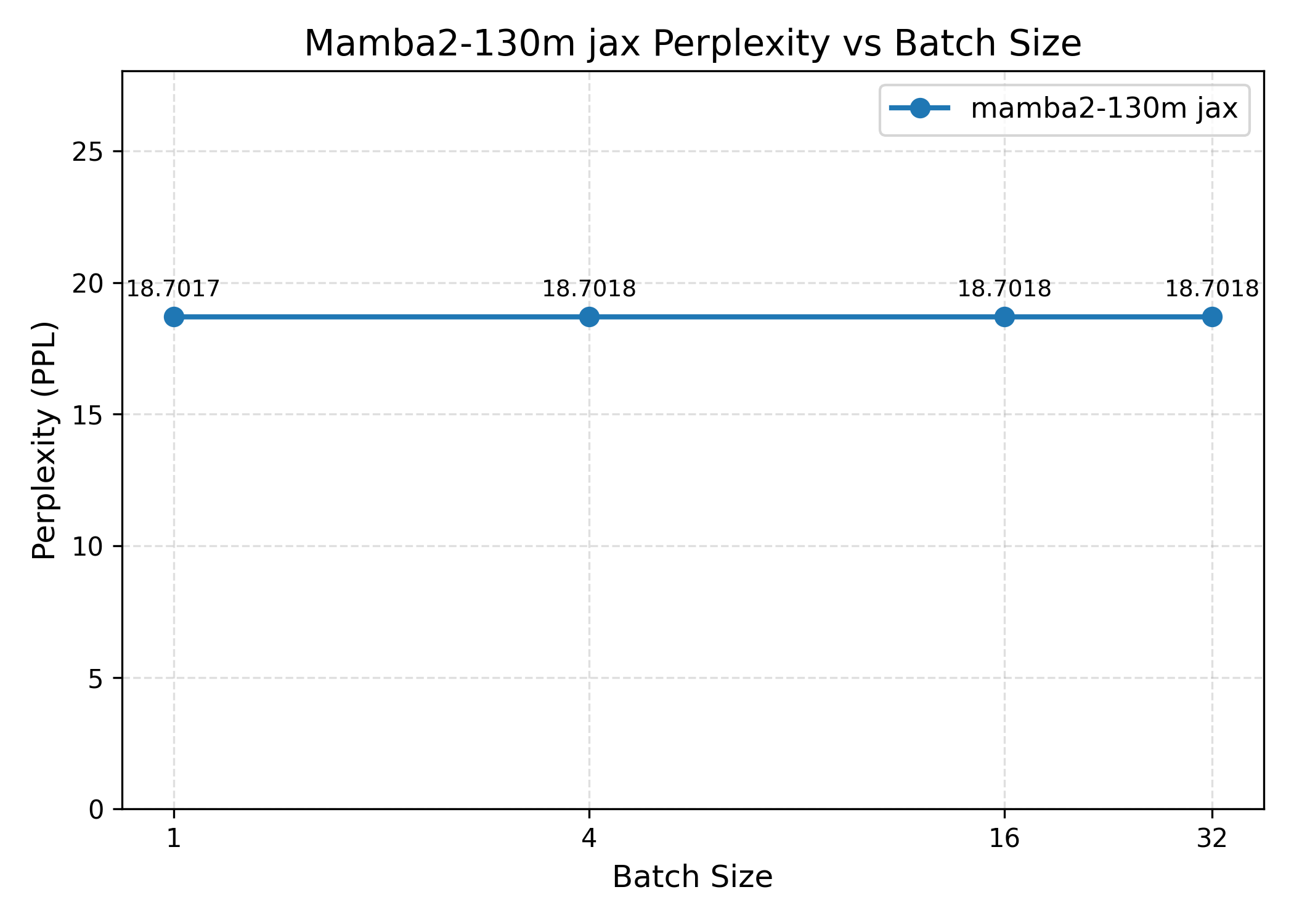}
\caption{WikiText-103 validation perplexity versus batch size for the $130$M model (JAX implementation). Perplexity is invariant to batch size for this checkpoint.}
\label{fig:ppl_batch}
\end{figure}

WikiText-103~\citep{merity2017pointer} validation perplexity is measured with stride $512$ against the Triton reference \texttt{mamba\_ssm} v$2.2.2$ under matched conditions, using float32 throughout, TF32 disabled, greedy decoding, and batch size $1$. Both implementations load the same five HuggingFace checkpoints. The largest absolute difference is $0.0005$, and the JAX implementation is invariant to batch size (Figure~\ref{fig:ppl_batch}).

\begin{table}[htbp]
\centering
\begin{tabular}{lccc}
\toprule
\textbf{Model} & \textbf{Triton PPL} & \textbf{JAX PPL} & \textbf{|$\Delta$|} \\
\midrule
130M & 18.7023 & 18.7019 & 0.0004 \\
370M & 13.1247 & 13.1244 & 0.0003 \\
780M & 10.8892 & 10.8886 & 0.0005 \\
1.3B & 9.5708 & 9.5704 & 0.0004 \\
2.7B & 8.3252 & 8.3250 & 0.0002 \\
\bottomrule
\end{tabular}
\caption{WikiText-103 validation perplexity (stride 512) for Triton reference
(\texttt{mamba\_ssm} v2.2.2) and JAX implementation. |$\Delta$| is the absolute difference between the JAX implementation and the Triton reference;
values within $\pm$0.0005 indicate functional equivalence.}
\label{tab:perplexity}
\end{table}

\subsection{Numerical Correctness}

Float32 addition is non-associative, and differing reduction orders between XLA and Triton produce a small accumulated absolute drift (${\sim}2 \times 10^{-4}$) through the $24$ residual layers. Perplexity remains unchanged at the reported precision, and the element-wise tolerances stay at float32 rounding scale.

\begin{table}[htbp]
\centering
\begin{tabular}{lll}
\toprule
\textbf{Output} & \textbf{Relative Tolerance} & \textbf{Absolute Tolerance} \\
\midrule
Last hidden state & $1 \times 10^{-5}$ & $1 \times 10^{-4}$ \\
Logits (first 256) & $1 \times 10^{-5}$ & $2 \times 10^{-4}$ \\
\bottomrule
\end{tabular}
\caption{Numerical parity against the PyTorch reference (\texttt{mamba\_ssm}, 130M
checkpoint, 512 tokens). Tolerances apply element-wise.}
\label{tab:numerical_parity}
\end{table}

Reference outputs are generated on GPU using the official \texttt{mamba\_ssm} PyTorch/CUDA package (v$2.2.2$) with float32 throughout and TF32 disabled, against the JAX implementation with default matmul precision set to its highest mode.

\subsection{JIT Compilation Cost}
\label{sec:jit_cost}

XLA compilation is a one-time cost for repeated inference calls and a recurring cost during interactive development. At the $2.7$B scale, the decode path takes $43$~seconds to compile at sequence length $4096$ (Table~\ref{tab:jit_time}).

\subsection{Ablations}
\label{sec:ablations}

Mask placement and decay precision are ablated independently. The compiled-loop comparison is reported with the throughput results in Table~\ref{tab:host_vs_scan}.

\paragraph{Mask placement.}

The segment-sum routine applies the lower-triangular causal mask to a precomputed matrix, which XLA folds into the surrounding fusion chain of prefix sum, subtraction, mask, and exponentiation. The ablated variant applies the same mask row by row inside a runtime loop over the $256$ chunk positions, using dynamic slice and update primitives. Output is bitwise identical, and the throughput cost is $82.8\%$, attributable to the fusion chain breaking at the loop boundary.

\begin{table}[htbp]
\centering
\begin{tabular}{lcc}
\toprule
\textbf{Masking Strategy} & \textbf{Prefill (tokens/s)} & \textbf{Output} \\
\midrule
Static mask (\texttt{jnp.tril}) & 42{,}631 & \multirow{2}{*}{Bitwise identical} \\
Dynamic row-wise mask (\texttt{fori\_loop}) & 7{,}330 \;($-82.8\%$) & \\
\bottomrule
\end{tabular}
\caption{Masking ablation on TPU v6e (1.3B model, BF16, prompt length 1024). Both
produce bitwise identical output; the dynamic variant breaks XLA's fusion chain.}
\label{tab:ablation_masking}
\end{table}

\paragraph{Decay precision.}

BF16 truncation of $\bar{A} = \exp(A_{\log})$ accumulates through the $24$-layer stack. The maximum absolute logit error reaches $0.013$, large enough to shift the output distribution. Upcasting to float32 for the exponentiation costs no measurable runtime and is required for correctness.

\begin{table}[htbp]
\centering
\begin{tabular}{lc}
\toprule
\textbf{Decay Dtype} & \textbf{Max Absolute Error (Logits)} \\
\midrule
\texttt{float32} (baseline) & 0.0 \\
\texttt{bfloat16} & 0.013 \\
\bottomrule
\end{tabular}
\caption{Decay precision ablation on TPU v6e (130M checkpoint, 24 layers, BF16,
prompt length 1024).}
\label{tab:ablation_precision}
\end{table}

\section{Discussion}
\label{sec:discussion}

On TPU v6e, cached decode reaches $64\%$ HBU because each step reads and writes a fixed-size SSM and convolution cache that XLA tiles into a stable memory-traffic pattern. Prefill reaches $15\%$ MFU at batch $1$ because the arithmetic intensity remains below the ${\sim}574$ FLOPs per byte needed to saturate v6e compute. Removing each implementation choice has a measurable cost. Row-wise runtime masking reduces prefill throughput by $82.8\%$ (Table~\ref{tab:ablation_masking}). A host-driven decode loop is $2.4\times$ slower than a compiled loop at $130$M (Table~\ref{tab:host_vs_scan}). Bfloat16 decay exponentiation introduces a $0.013$ maximum absolute logit error (Table~\ref{tab:ablation_precision}). Keeping the SSM cache as a JAX PyTree avoids host round-trips by carrying the $O(1)$ state through the compiled loop.

\section{Limitations}
\label{sec:limitations}

\textbf{Profiling scope.} MFU and HBU are reported on TPU v6e and NVIDIA L40S. Other XLA backends (TPU v4 and v5e, CPU, AMD GPUs via OpenXLA) have different fusion strategies and scheduling; absolute utilisation will vary.

\textbf{Fixed chunk size.} All experiments use $L = 256$, the default of \citet{dao2024transformers}. Chunk size is a tuning variable orthogonal to the compiler-first pattern; its interaction with hardware tiling is unmeasured here.

\textbf{Inference batch policies.} Inference numbers are reported at fixed batch sizes. Continuous batching and dynamic memory paging~\citep{kwon2023pagedattention} are scheduling concerns not implemented here; the cache primitive is compatible with such schedulers.

\textbf{Training regime.} The implementation targets inference and trains at numerical parity, but the reduced L40S comparison is favourable only for small models and short horizons. Forward$+$backward time is up to $2.8\times$ lower than the Triton reference at the $130$M, $512$-token point, and the advantage shrinks as model size and sequence length grow. Many time-series~\citep{rangapuram2018deepstate, wang2024smamba}, control, and scientific sequence workloads use sub-$100$M models, below the smallest checkpoint measured here. At larger scale, the relationship crosses over. Beyond roughly $780$M parameters or $2048$ tokens the JAX step is several times slower, and the $1.3$B (sequence length $\ge 4096$) and $2.7$B models exceed L40S memory under the JAX path. Matching the Triton reference at LLM scale would require kernel-level work, such as a custom Pallas or Triton fused backward, outside the compiler-first path evaluated here.

\textbf{Compiler-hostile primitives.} Data-dependent memory access (gather and scatter over runtime indices), warp-level synchronisation, and data-dependent control flow are not exposed through the standard JAX primitive set. SSD does not require any of them. Architectures that do require them do not satisfy the structural conditions analyzed here.

\textbf{Compiler maturity.} A newly bootstrapped XLA backend may not immediately match the fusion and tiling quality of the mature TPU and GPU backends; the numbers reported here should not be extrapolated to early-stage backends without measurement.

\textbf{Compilation cost.} The compiler-first path trades one-time XLA compilation for hardware portability. At the $2.7$B scale the decode path takes $43$~seconds to compile at sequence length $4096$, which dominates iterative-research wall-clock and is amortised only across many inference calls.

\section{Conclusion}
\label{sec:conclusion}

The four structural conditions of state space duality (a diagonal state matrix, a chunkable recurrence, einsum-dominated compute, and static control flow) are sufficient for XLA to produce competitive code for Mamba-2 inference without custom kernels. From a single source, the implementation reaches $15\%$ model FLOP utilisation prefill and $64\%$ hardware bandwidth utilisation decode on TPU v6e, both at the batch-$1$ roofline ceilings; reproduces sequence-length-independent cached decode on NVIDIA L40S; and matches WikiText-103 perplexity within $\pm 0.0005$ points across all five model scales.

\section*{Author Contributions}

Cosmo Santoni led the core architecture design, formulating the compiler-first state space duality implementation and the $O(1)$ autoregressive caching mechanism, and conducted the TPU v6e evaluations. Anmol Thapar led the cross-hardware evaluation, including the NVIDIA L40S inference measurements, the reduced training-step comparison, and the downstream perplexity validation. Both authors contributed to experimental design, ablation studies, and manuscript revision.

\section*{Acknowledgements}

We thank Jiyoun Ha, James Chapman, and Skye Wanderman-Milne at Google for code review,
testing and validation strategy, and technical guidance during development and
integration of the Mamba-2 module into Bonsai, and Carlos Araya at Google for
facilitating the collaboration. Timothy Hitge at Imperial College London assisted with
experiment scripting and GPU result collection. This research was supported in part
with Cloud TPUs from Google's TPU Research Cloud (TRC).

\appendix

\section{Experimental Details}
\label{sec:experimental_details}

\subsection{Supported Checkpoints}

All five model sizes (130M--2.7B) are loaded from the original HuggingFace weights
(\texttt{state-spaces/mamba2-*}).

\subsection{Benchmark Configurations}
\label{sec:benchmark_config}

\textbf{Decode sweep (single-stream, TPU v6e).} $5$ models $\times$ $6$ sequence lengths ($128$--$4096$) $\times$ $3$ methods (cached scan, cached host, non-cached) $\times$ $5$ timed runs after JIT warm-up. Prompt length fixed at $16$ tokens.

\textbf{Decode sweep (single-stream, L40S).} Same protocol as TPU v6e above.

\textbf{Prefill sweep (TPU v6e).} $5$ models $\times$ $3$ prompt lengths ($1024$, $4096$, $8192$) $\times$ $5$ timed runs. XLA cost analysis extracted per configuration.

\textbf{Training-step sweep (L40S).} $3$ checkpoints ($130$M--$780$M) $\times$ batch size $1$ $\times$ $3$ sequence lengths $\{512, 1024, 2048\}$ $\times$ $10$ timed steps after $10$ warm-up steps. Each timed cell is a forward $+$ backward pass relative to the Triton reference. The $1.3$B and $2.7$B checkpoints, and sequence lengths beyond $2048$, exceed L40S memory under the JAX path and are omitted.

\textbf{Perplexity.} WikiText-103 validation split, stride $512$, batch size $1$, float32 throughout, TF32 disabled, for both the JAX implementation and \texttt{mamba\_ssm} v$2.2.2$.

\subsection{Reproducibility}
\label{sec:reproducibility}

\begin{table}[htbp]
\centering
\begin{tabular}{ll}
\toprule
\textbf{Component} & \textbf{Version / Value} \\
\midrule
JAX & 0.9.0 \\
jaxlib & 0.9.0.1 \\
XLA / libtpu (TPU runtime) & Bundled with jaxlib 0.9.0.1 \\
Python & 3.12 \\
Flax (NNX) & 0.12.4 \\
PyTorch (golden outputs) & 2.10.0 \\
\texttt{mamba\_ssm} (golden outputs / perplexity) & 2.2.2 \\
\midrule
Bonsai core (PR \#103) & \texttt{a907b75} \\
Bonsai caching (PR \#131) & \texttt{d8f8d11} \\
\midrule
HuggingFace checkpoint IDs & \texttt{state-spaces/mamba2-130m} \\
 & \texttt{state-spaces/mamba2-370m} \\
 & \texttt{state-spaces/mamba2-780m} \\
 & \texttt{state-spaces/mamba2-1.3b} \\
 & \texttt{state-spaces/mamba2-2.7b} \\
\midrule
\texttt{jax\_default\_matmul\_precision} & \texttt{"highest"} (correctness); default
(throughput) \\
\texttt{torch.backends.cuda.matmul.allow\_tf32} & \texttt{False} (golden outputs and
perplexity) \\
Chunk size $L$ & 256 \\
BF16 compute dtype & All throughput and ablation runs \\
\bottomrule
\end{tabular}
\caption{Software versions and configuration flags.}
\label{tab:reproducibility}
\end{table}

Exact software versions and configuration flags are listed in Table~\ref{tab:reproducibility}. Benchmark scripts, configuration files, and reproduction instructions are in the
project repository~\citep{mamba2jax}.

\section{Additional Results}
\label{sec:additional_results}

\subsection{Full Single-Stream Decode Throughput (TPU v6e)}
\label{sec:full_decode_throughput}

Cached throughput is sequence-length independent, whereas the non-cached path slows sharply with sequence length (Table~\ref{tab:tpu_throughput}).

\begin{table}[htbp]
\centering
\begin{tabular}{llcccccc}
\toprule
& & \multicolumn{6}{c}{\textbf{Throughput (Tokens/Second) by Sequence Length}} \\
\cmidrule(lr){3-8}
\textbf{Model} & \textbf{Method} & \textbf{128} & \textbf{256} & \textbf{512} &
\textbf{1024} & \textbf{2048} & \textbf{4096} \\
\midrule
130M & Cached   & 1588 & 1609 & 1627 & 1635 & 1639 & 1641 \\
     & Non-Cached & 903 & 898 & 626 & 278 & 132 & 56 \\
\midrule
370M & Cached   & 626 & 634 & 639 & 641 & 641 & 641 \\
     & Non-Cached & 495 & 348 & 205 & 124 & 39 & 18 \\
\midrule
780M & Cached   & 318 & 321 & 322 & 322 & 323 & 323 \\
     & Non-Cached & 311 & 229 & 129 & 60 & 24 & 9 \\
\midrule
1.3B & Cached   & 188 & 189 & 190 & 190 & 190 & 190 \\
     & Non-Cached & 185 & 99 & 66 & 32 & 13 & 7 \\
\midrule
2.7B & Cached   & 94 & 94 & 95 & 95 & 95 & 95 \\
     & Non-Cached & 95 & 63 & 40 & 17 & 8 & 3 \\
\bottomrule
\end{tabular}
\caption{Autoregressive decoding throughput on TPU v6e (batch size 1, \texttt{fori\_loop}
scan path).}
\label{tab:tpu_throughput}
\end{table}

\subsection{Full Peak Memory During Autoregressive Generation (TPU v6e)}
\label{sec:full_memory_tables}

The cached path keeps peak device memory constant, whereas the non-cached path grows with sequence length (Table~\ref{tab:peak_memory}).

\begin{table}[htbp]
\centering
\begin{tabular}{llccccccc}
\toprule
& & \multicolumn{6}{c}{\textbf{Peak Memory (MB) by Sequence Length}} \\
\cmidrule(lr){3-8}
\textbf{Model} & \textbf{Method} & \textbf{128} & \textbf{256} & \textbf{512} &
\textbf{1024} & \textbf{2048} & \textbf{4096} \\
\midrule
130M & Cached     & 545.6 & 545.6 & 545.6 & 545.6 & 545.6 & 545.6 \\
     & Non-Cached & 565   & 585   & 624   & 702   & 857   & 1169 \\
\midrule
370M & Cached     & 1591.9 & 1591.9 & 1591.9 & 1591.9 & 1591.9 & 1591.9 \\
     & Non-Cached & 1644   & 1696   & 1799   & 2007   & 2422   & 3251 \\
\midrule
780M & Cached     & 3322.8 & 3322.8 & 3322.8 & 3322.8 & 3322.8 & 3322.8 \\
     & Non-Cached & 3401   & 3478   & 3634   & 3945   & 4566   & 5809 \\
\midrule
1.3B & Cached     & 5620.2 & 5620.2 & 5620.2 & 5620.2 & 5620.2 & 5620.2 \\
     & Non-Cached & 5724   & 5827   & 6035   & 6450   & 7279   & 8938 \\
\midrule
2.7B & Cached     & 10861.8 & 10861.8 & 10861.8 & 10861.8 & 10861.8 & 10861.8 \\
     & Non-Cached & 11035   & 11208   & 11553   & 12244   & 13627   & 16392 \\
\bottomrule
\end{tabular}
\caption{Peak memory during autoregressive generation on TPU v6e (batch size 1).}
\label{tab:peak_memory}
\end{table}

\subsection{Hardware Utilisation Summary (TPU v6e)}
\label{sec:utilisation_summary}

\begin{figure}[htbp]
\centering
\includegraphics[width=0.75\textwidth]{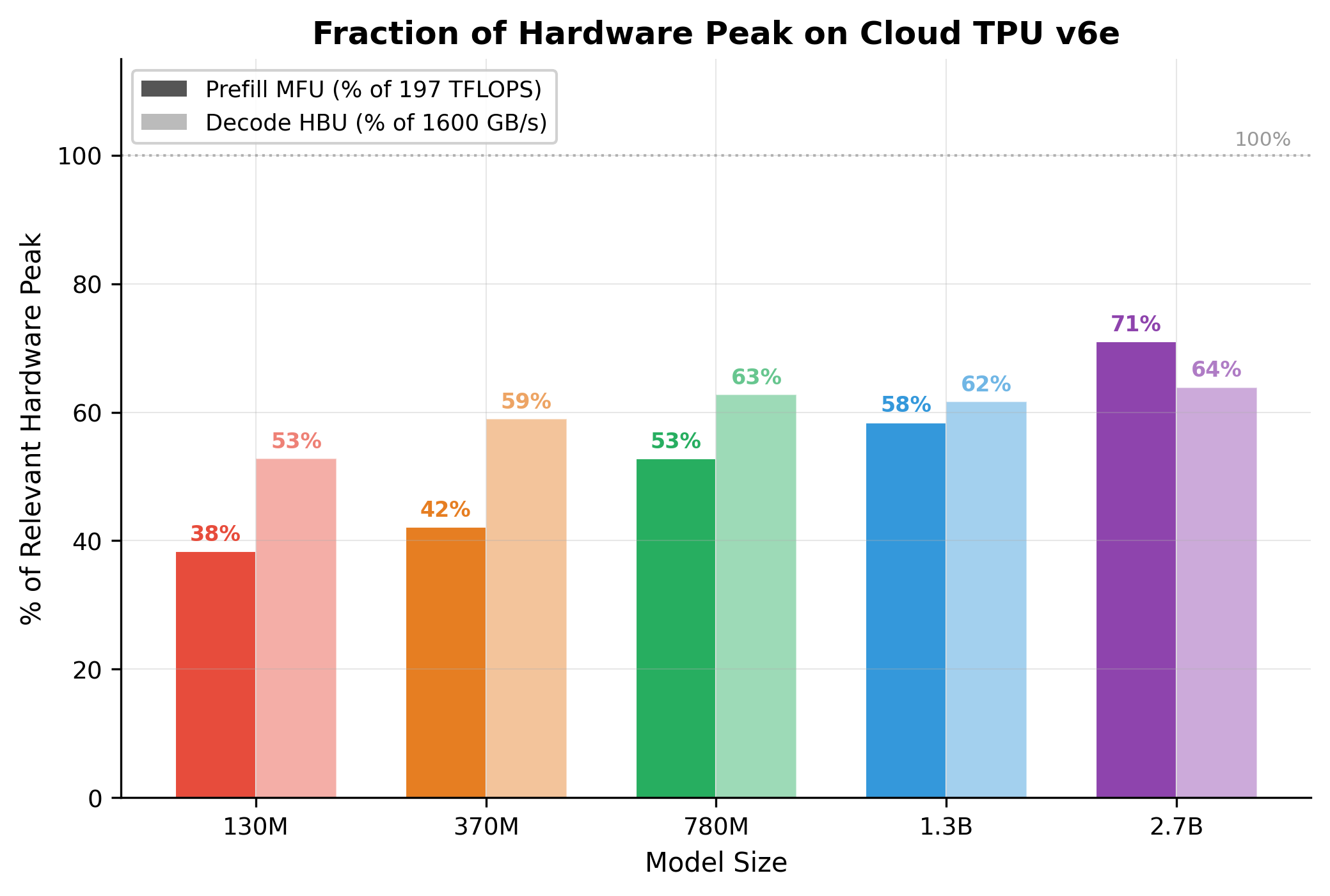}
\caption{Fraction of hardware peak on Cloud TPU v6e (batch size $1$). Solid bars give best prefill MFU (\% of $918$~TFLOPS); faded bars give mean decode HBU (\% of $1600$~GB/s). Utilisation increases with model size in both regimes.}
\label{fig:device_ceilings}
\end{figure}

Prefill MFU and mean decode HBU both increase with model size across the measured TPU v6e regimes (Figure~\ref{fig:device_ceilings}).

\subsection{JIT Compilation Details}
\label{sec:jit_compilation_details}

One-time XLA compilation time grows with model size and decode horizon (Table~\ref{tab:jit_time}).

\begin{table}[htbp]
\centering
\begin{tabular}{lccc}
\toprule
& \multicolumn{3}{c}{\textbf{JIT Compilation Time (seconds)}} \\
\cmidrule(lr){2-4}
\textbf{Model} & \textbf{Prefill (1024)} & \textbf{Decode (128)} & \textbf{Decode
(4096)} \\
\midrule
130M & 5.5 & 5.6 & 2.5 \\
370M & 10.2 & 13.0 & 6.4 \\
780M & 13.0 & 13.7 & 12.6 \\
1.3B & 10.2 & 14.9 & 21.4 \\
2.7B & 15.8 & 19.5 & 43.0 \\
\bottomrule
\end{tabular}
\caption{XLA JIT compilation time on TPU v6e. One-time costs; subsequent calls reuse
the compiled program.}
\label{tab:jit_time}
\end{table}

\subsection{Reduced Training-Step Comparison (NVIDIA L40S)}
\label{sec:training_step_appendix}

The reduced training-step comparison uses a single NVIDIA L40S, the compiler-first
JAX path, and the Triton reference (\texttt{mamba\_ssm} v$2.2.2$), at batch size $1$ for
the three smallest checkpoints and sequence lengths $\{512, 1024, 2048\}$. Each cell is
the mean forward$+$backward time over ten timed steps after ten warm-ups. The optimiser
update is excluded because its JAX timing was dominated by a measurement artefact rather
than a reliable steady-state cost. At $130$M and $512$ tokens, the JAX path is $64.8\%$
faster; the advantage shrinks with model size and sequence length and becomes a slowdown
by $2048$ tokens for every measured checkpoint. The $1.3$B model exceeds L40S memory at
sequence lengths $\ge 4096$, and the $2.7$B model exceeds memory at every measured
sequence length under the JAX path.

\begin{table}[htbp]
\centering
\begin{tabular}{llccc}
\toprule
\textbf{Model} & \textbf{Seq.\ length} & \textbf{JAX (ms)} & \textbf{Triton (ms)} & \textbf{$\Delta\%$} \\
\midrule
\multirow{3}{*}{130M} & 512  & 25.9  & 73.7  & $-64.8$ \\
                      & 1024 & 45.2  & 72.4  & $-37.5$ \\
                      & 2048 & 86.7  & 68.0  & $+27.6$ \\
\midrule
\multirow{3}{*}{370M} & 512  & 62.8  & 147.0 & $-57.3$ \\
                      & 1024 & 115.8 & 128.6 & $-9.9$ \\
                      & 2048 & 229.6 & 151.4 & $+51.7$ \\
\midrule
\multirow{3}{*}{780M} & 512  & 104.5 & 148.2 & $-29.5$ \\
                      & 1024 & 316.3 & 136.3 & $+132.1$ \\
                      & 2048 & 572.9 & 148.0 & $+287.1$ \\
\bottomrule
\end{tabular}
\caption{Reduced training-step comparison on a single NVIDIA L40S. Cells give mean
forward$+$backward time (ms) over ten timed steps for the JAX compiler-first path and the
Triton reference (\texttt{mamba\_ssm} v$2.2.2$), batch size $1$. $\Delta =
(t_{\text{JAX}} - t_{\text{Triton}})/t_{\text{Triton}}$; negative means the JAX path is
faster. The JAX path is faster for small models at short sequences and crosses over to
several times slower as model size and sequence length grow. The $1.3$B model (sequence
length $\ge 4096$) and the $2.7$B model (all sequence lengths) exceed L40S memory under
the JAX path and are omitted.}
\label{tab:training_step_appendix}
\end{table}

\section{Exact SSD Einsum Signatures}
\label{sec:einsum_signatures}

Axis labels: \texttt{b}=batch, \texttt{c}=chunk, \texttt{l}/\texttt{s}=sequence-within-chunk,
\texttt{h}=head, \texttt{n}=state, \texttt{p}=head\_dim, \texttt{z}=target chunk.

\begin{verbatim}
# Intra-chunk output Y_diag
Y = jnp.einsum('bclhn,bcshn,bhcls,bcshp->bclhp', C, B, L, X)

# State accumulation (per-chunk hidden states)
states = jnp.einsum('bclhn,bhcl,bclhp->bchpn', B, decay, X)

# Inter-chunk recurrence (scan update)
new_states = jnp.einsum('bhzc,bchpn->bzhpn', decay_chunk, states)
\end{verbatim}

\bibliographystyle{plainnat}
\bibliography{refs}

\end{document}